\pdfoutput=1
\documentclass[11pt]{article}

\usepackage[]{acl}

\usepackage{times}
\usepackage{latexsym}

\usepackage[T1]{fontenc}
\usepackage[utf8]{inputenc}

\usepackage{microtype}

\usepackage{times}
\usepackage{latexsym}
\usepackage{hyperref}
\usepackage{color, colortbl}
\definecolor{Gray}{gray}{0.9}
\usepackage{array}
\usepackage{booktabs}
\usepackage{multirow}
\usepackage{amsmath}
\usepackage{amssymb}
\usepackage{graphicx}
\usepackage{enumerate}
\usepackage{diagbox}
\usepackage{mathtools}
\usepackage{paralist}
\usepackage{tabularx}
\usepackage{lipsum}
\usepackage{ragged2e}
\usepackage{pifont}

\usepackage{microtype}

\usepackage{inconsolata}
\usepackage{url}

\newcommand{\cmark}{\ding{51}}%
\newcommand{\xmark}{\ding{55}}%

\definecolor{entitycolor}{HTML}{018161}
\definecolor{relation1color}{HTML}{d95f02}
\definecolor{relation2color}{HTML}{6560a3}

\newcommand*{\affaddr}[1]{#1} % No op here. Customize it for different styles.
\newcommand*{\affmark}[1][*]
{\textsuperscript{#1}}

\newcommand{\tabincell}[2]{\begin{tabular}{@{}#1@{}}#2\end{tabular}} 

\newcommand{\dataname}{{{\textsc{Com$^2$}}}}
\newcommand{\slashn}{\textbackslash n}

\usepackage{cleveref}
\crefformat{section}{\S#2#1#3}
\crefformat{subsection}{\S#2#1#3}
\crefformat{subsubsection}{\S#2#1#3}
\crefrangeformat{section}{\S#3#1#4 to~\S#5#2#6}
\crefmultiformat{section}{\S#2#1#3}{ and~\S#2#1#3}{, #2#1#3}{ and~#2#1#3}
\Crefformat{figure}{#2Fig.~#1#3}
\Crefmultiformat{figure}{Figs.~#2#1#3}{ and~#2#1#3}{, #2#1#3}{ and~#2#1#3}
\Crefformat{table}{#2Tab.~#1#3}
\Crefmultiformat{table}{Tabs.~#2#1#3}{ and~#2#1#3}{, #2#1#3}{ and~#2#1#3}
\Crefformat{appendix}{#2Appx.~\S#1#3}

\newif\ifcomments
\commentstrue
\ifcomments
\usepackage[normalem]{ulem}
\definecolor{ABpurple}{rgb}{0.8,0.0,0.8}
\newcommand\ab[1]{\textcolor{ABpurple}{\textsf{\scriptsize[\textbf{AB\@:} #1]}}} % inline comment
\newcommand\abi[1]{\textcolor{ABpurple}{#1}} %text change
\newcommand\abm[1]{\marginpar{\raggedright\tiny\textcolor{ABpurple}{\textsf{{\bfseries AB\@:} #1}}}} %margin comment
\newcommand\abs{\bgroup\markoverwith{\textcolor{ABpurple}{\rule[.4ex]{2pt}{0.8pt}}}\ULon} % strike
\else
\newcommand\ab[1]{}
\newcommand\abi[1]{\ignorespaces}
\newcommand\abm[1]{}
\newcommand\abs[1]{#1}
\fi

\title{Complex Reasoning over Logical Queries on Commonsense Knowledge Graphs}

\author{
Tianqing Fang\affmark[1,2]\thanks{\quad Work done during internship at EPFL.}~,
Zeming Chen\affmark[2],
Yangqiu Song\affmark[1],
Antoine Bosselut\affmark[2]\\
\affaddr{\affmark[1]CSE, HKUST, Hong Kong SAR, China} \\
\affaddr{\affmark[2]NLP Lab, IC, EPFL, Switzerland}\\
\texttt{\{tfangaa, yqsong\}@cse.ust.hk, \{zeming.chen, antoine.bosselut\}@epfl.ch} \\ 
}

\begin{document}
\maketitle
\begin{abstract}

% Complex commonsense reasoning on narratives requires the ability to reason regarding multiple events and their relationships, as well as infer hidden context that is not explicitly stated.
Event commonsense reasoning requires the ability to reason about the relationship between events, as well as infer implicit context underlying that relationship. %, which state-of-the-art language models still struggle to perform.
% However, data scarcity has always been an issue for conducting narrative-level inferences.
However, data scarcity makes it challenging for language models to learn to generate commonsense inferences 
% for complex contexts and questions.
for contexts and questions involving interactions between complex events.
To address this demand, we present \dataname{} (\textbf{COM}plex \textbf{COM}monsense), a new dataset created by sampling multi-hop logical queries (e.g., \textit{the joint effect or cause of both event A and B, or the effect of the effect of event C}) from an existing commonsense knowledge graph (CSKG), and verbalizing them using handcrafted rules and large language models into multiple-choice and text generation questions.

% To facilitate evaluation and advancement in Complex Commonsense Reasoning, 

% In \dataname, we curate a human-annotated evaluation set containing 1.3K context-inference pairs presented through multiple-choice question answering and text generation tasks.
% In addition, the verbalized complex queries can automatically form a distantly supervised corpus. 
Our experiments show that 
% state-of-the-art language models still struggle in 
% performing complex commonsense reasoning,
% answering multi-hop logical queries in CSKGs.
language models trained on \dataname{} exhibit significant improvements in complex reasoning ability, resulting in enhanced
zero-shot performance in both in-domain and out-of-domain tasks for question answering and generative commonsense reasoning, without expensive human annotations.\footnote{Code and data are available at \url{https://github.com/tqfang/complex-commonsense-reasoning}}

\end{abstract}

\section{Introduction}

Large language models struggle to effectively perform reasoning when presented with complex tasks, such as reasoning about multiple events and their relationships. %, as well as inferring implicit context to facilitate subsequent reasoning.
This shortcoming is due to both the inherent difficulty of reasoning over multiple pieces of information, as well as a lack of adequate-scale, supervised training datasets for learning~\cite{DBLP:conf/acm/ZhaoGLYM023}.
Unfortunately, complex and multi-hop commonsense reasoning benchmarks~\cite{paracomet} are both technically challenging and financially expensive to curate. Consequently, previous efforts either constructed datasets (a) with simpler reasoning structures, such as single-hop chains~\cite{DBLP:conf/emnlp/MostafazadehKMB20}, (b) using distant supervision based on one-hop inference~\cite{paracomet}, or (c) with human-annotations, but at a relatively small scale~\cite{comet-m}.

\begin{figure}[t]
    \centering
    \includegraphics[width=1\linewidth]{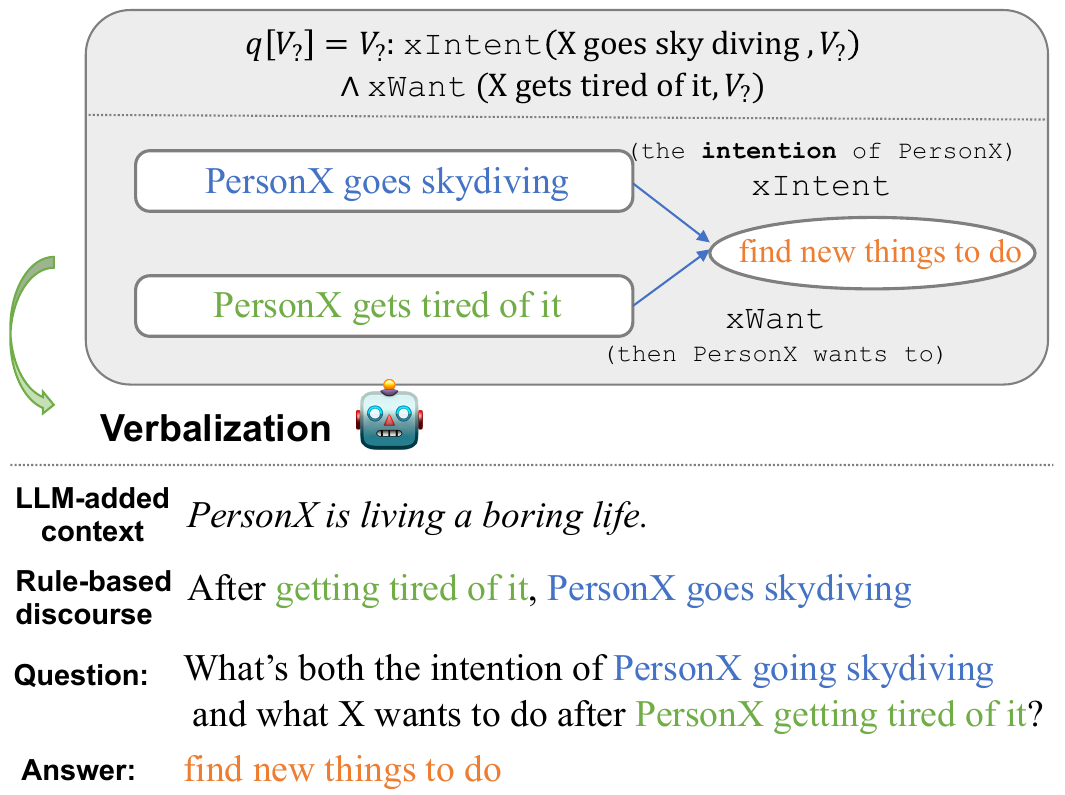}
    % \vspace{-0.1in}
    \caption{An example of conjunctive logical queries and their verbalization as complex commonsense inferences. }
    \label{fig:Introduction_demo}
    % \vspace{-0.2in}
\end{figure}

To alleviate this training data bottleneck, recent works have explored extracting and formulating questions from existing CommonSense Knowledge Graphs (CSKGs; \citealp{DBLP:conf/aaai/HwangBBDSBC21}), which store commonsense triples. However, using CSKGs to produce high-quality reasoning datasets poses several challenges. First, while the shared entities in commonsense triples encode a complex, interconnected graph structure, the sparsity of this structure limits the number of potential questions that encode more than one reasoning hop \citep{DBLP:conf/emnlp/SapRCBC19,DBLP:conf/emnlp/0002HJWLYZ0AKS023}. Second, triples in CSKGs are represented in a context-free manner, such as the event ``PersonX gets tired of it'' in \Cref{fig:Introduction_demo}, yielding ambiguous (and sometimes incorrect) human annotations in the CSKG, e.g., ATOMIC \citep{DBLP:conf/aaai/SapBABLRRSC19} has an error rate of over 10\%. These errors propagate when triples are naively combined to construct reasoning questions. Finally, also because triples in CSKGs are represented in a context-free manner, additional context must be added to make questions fluent, a problem exacerbated in multi-hop settings where the entities of multiple reasoning hops must be coherently verbalized together.

In this paper, we construct \dataname{} (\textbf{COM}plex \textbf{COM}monsense), a novel commonsense reasoning dataset using multi-hop queries in commonsense knowledge graphs to construct question answer pairs requiring complex narrative reasoning. 
To build this dataset, we use \textit{conjunctive logical queries}~\cite{DBLP:conf/nips/HamiltonBZJL18}, a subset of First-Order Logical queries that use existential quantifiers and conjunction.
% Such logical queries are natural sources for building narratives grounded with complex commonsense causes or effects.
% The anchor entities within complex logical queries serve as natural sources for deriving multi-event narratives, where the predicates themselves formulate intricate reasoning questions, including common causes, common effects, and abduction.
The multi-hop projection operation involves inferring hidden contexts, while the intersection operation enables reasoning among multiple events, encompassing common cause or effect, and abduction.
For example, in \Cref{fig:Introduction_demo}, an intersection of two triples can be verbalized to a short narrative, and the process of inferring the common tail can be seen as an \textit{abduction} of the hidden cause between the two heads. 

To address the challenges above, we propose to first \textit{densify} the CSKG to merge nodes with high semantic similarity, increasing the connectivity of the graph. Then, we use an off-the-shelf plausibility scorer to filter out low quality triples, avoiding error propagation as we construct more complicated queries. Finally, we verbalize the queries to a natural language context with handcrafted rules and Large Language Models to derive coherent and informative narrative contexts for our questions. Our final \dataname{} dataset comprises 790K question-answer pairs (both with multiple-choice and generative answer settings), including 1.3K examples that we manually verify for evaluation. 

Our results demonstrate the challenges faced by even powerful LLMs and supervised question answering models on the \dataname{} dataset, underscoring the difficulty of performing complex multi-hop reasoning.
Moreover, fine-tuning question answering models and generative commonsense inference models on \dataname{} leads to substantial improvements across eight commonsense reasoning datasets, showing the efficacy of our framework for boosting commonsense reasoning ability.

To conclude, our contributions are three-fold. First, we present a pipeline for sampling and verbalizing complex logical queries from CSKGs, to form a complex commonsense reasoning benchmark, \dataname{}, with minimal human effort.
Second, we benchmark the complex reasoning ability of various state-of-the-art language models and question answering models on \dataname{}.
% and show that models still fall short in such complex reasoning and perform poorly even for powerful GPT-4.
Finally, we validate the benefit of fine-tuning on \dataname{} on eight zero-shot commonsense reasoning datasets.

% on two datasets.
% as reflected by downstream evaluation in Para-COMET and COMET-M.

\section{Background and Related Work}

% In this paper, we narrow down the concept of complex reasoning to reasoning over multiple triples and leveraging implicit context. This aligns with the practice of employing complex logical queries over knowledge graphs.

\paragraph{Complex Logical Queries}
Recent years have witnessed significant progress in reasoning on one-hop relational data~\cite{transe, rotate, DBLP:journals/inffus/LinMLXC23}. 
In addition to one-hop reasoning, further works have explored handling complex logical structures, involving \textit{reasoning on unobserved edges and multiple entities and variables}~\cite{query2box, DBLP:conf/nips/WangYS21, DBLP:conf/iclr/WangSWS23, DBLP:journals/corr/abs-2305-19068}.
In this paper, we focus on conjunctive logical queries~\cite{DBLP:conf/nips/HamiltonBZJL18}, a subset of first-order logic that is defined with logical operators such as
existential quantifiers $\exists$ and conjunctions $\wedge$. 
Conjunctive logical queries require a set of anchor entities, $\mathcal{V}$, a unique target entity $V_?$ representing the answer to the query, and a set of existential quantified variables $V_1, \cdots, V_m$, and are defined as the conjunction of literals $e_1, \cdots, e_n$:

\vspace{-1em}
\begin{align}
    q & = V_?, \exists V_1, \cdots, V_m: e_1 \wedge e_2 \wedge \cdots  \wedge e_n
    % & \small{\text{where}\ e_i = r(v_j, V_k), V_k \in \{V_?, V_1, \cdots, V_m\},} \nonumber \\ 
    % & v_j \in \mathcal{V}, r \in \mathcal{R}, \nonumber \\
    % & 
\end{align}

\noindent where $e_i$ is an edge involving variable nodes and anchor nodes, satisfying $e_i = r(v_j, V_k), V_k \in \{V_?, V_1, \cdots, V_m\}$, $v_j \in \mathcal{V}, r \in \mathcal{R}$, or $e_i = r(V_j, V_k), V_j, V_k \in \{V_?, V_1, \cdots, V_m\}, j\neq k, r \in \mathcal{R}$. $\mathcal{R}$ is the set of relations defined in the KB.

% The example in \Cref{fig:Introduction_demo} is a simple conjunctive query involving two anchor variables, $V_1$ = ``PersonX has second thoughts'', and $V_2$ = ``PersonX throws in the towel'', and the answer variable $V_?$ = ``give up''. The query takes the form $q=V_? : \texttt{xEffect}(V_1, V_?) \wedge \texttt{xIntent}(V_2, V_?)$.

Previous efforts on answering logical queries on knowledge graphs focus on constructing box embeddings~\cite{query2box}, embeddings based on beta distributions~\cite{betaE}, particle simulations~\cite{DBLP:conf/naacl/BaiWZS22}, and computation tree optimization~\cite{DBLP:conf/icml/BaiLLH23}.
Other related works focus on leveraging two-hop projection and intersection queries in ConceptNet to improve commonsense question answering~\cite{DBLP:journals/corr/abs-2305-05936}, inferring missing entities in verbalized complex queries on factual knowledge graphs~\cite{ding2023knowledge}, and developing an LLM agent for complex operators within the KG~\cite{DBLP:journals/corr/abs-2402-11163}.
Instead of relying on embeddings or limited query types for matching synthetic logical queries, 
we leverage the concept of logical queries to effectively acquire complex reasoning data from CSKGs with minimum human efforts.

% Additionally, we incorporate a verbalization process to ensure that the queries are human-readable and genuinely valuable for downstream reasoning tasks.

\paragraph{Complex Commonsense Reasoning} Recent advances in commonsense reasoning have been driven by the construction of human-annotated \citep{DBLP:conf/aaai/SpeerCH17,DBLP:conf/aaai/SapBABLRRSC19,DBLP:conf/aaai/HwangBBDSBC21,DBLP:conf/naacl/JiangBBC21,DBLP:conf/emnlp/MostafazadehKMB20, krishna2017visual, shen2024vcd} and human-validated \citep{DBLP:conf/naacl/WestBHHJBLWC22,DBLP:conf/acl/GaoBOBKWMB23} CommonSense Knowledge Graphs (CSKG). 
A common approach to create challenges for commonsense reasoning involves constructing tasks in the form of question-answering~\cite{DBLP:conf/naacl/TalmorHLB19, DBLP:conf/emnlp/SapRCBC19}, knowledge base completion~\cite{DBLP:conf/aaai/MalaviyaBBC20, yang-etal-2023-end} and population~\cite{DBLP:conf/www/FangZWSH21, DBLP:conf/emnlp/FangWCHZSH21}, grounding~\cite{DBLP:conf/emnlp/GaoHKWMB22}, and daily dialogue~\cite{DBLP:conf/emnlp/0002HJWLYZ0AKS023}, based on CSKGs.
However, most of those previous benchmarks are based on one-hop triples.

In contrast, real-world situations in narratives usually involve more complicated reasoning across multiple events, sentences, and paragraphs~\cite{DBLP:conf/ijcai/SchenkA75}. 
Previous works learn representations of narrative chains~\cite{DBLP:conf/acl/ChambersJ08, DBLP:conf/eacl/PichottaM14} and draw inferences~\cite{DBLP:conf/acl/FangBV22, DBLP:conf/acl/YuanCFGSJXY23}.
To address more complex paragraph-level or multi-event reasoning, ParaCOMET~\cite{paracomet} proposed to pre-train on distantly supervised one-hop paragraph-level commonsense inferences, and COMET-M~\cite{comet-m} was fine-tuned on a crowdsourced corpus focusing on reasoning on multiple events.
Instead of crowdsourcing or using language models to distill complex inferences, we provide narrative-level inference by verbalizing complex logical queries over CSKGs, to effectively acquire grounded inferences at scale. 
% Moreover, the reasoning question in previous works are also usually one-hop, e.g., asking about the causes of a whole sentence of asking the effect of each individual event triggers.
% In our work, we study the situations where the reasoning question involves multiple relations, e.g., the joint causes of several events.
% Moreover, in previous works, reasoning questions also typically focused on one-hop scenarios, like determining the causes of a complete sentence or the effects of individual event triggers. In contrast, our work explores reasoning questions involving multiple relations, such as identifying the joint causes of several events.

% \begin{figure}[t]
%     \centering
%     \includegraphics[width=1.0\linewidth]{image/overview.pdf}
%     % \vspace{-0.12in}
%     \caption{Overview of the construction process. }
%     \label{fig:query_types}
%     \vspace{-0.2in}
% \end{figure}

\begin{figure}[t]
    \centering
    \includegraphics[width=1.0\linewidth]{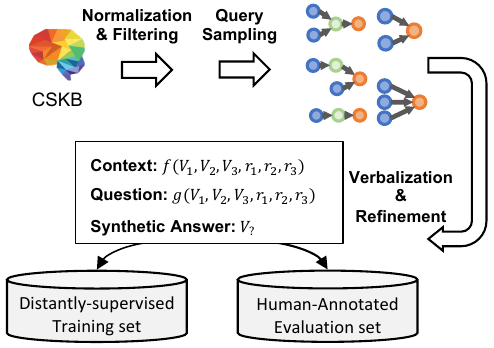}
    \caption{Overview of the construction process. $f$ represents a verbalization function for the context, and $g$ represents the one for the question. }
    \label{fig:overview}
\end{figure}

\section{Methodology}\label{sec:methodology}

In this section, we introduce the construction details of \dataname{}, including the pre-processing, sampling, and verbalization of complex queries, as well as the details of human annotations. An ovewview of the pipeline is presented in \Cref{fig:overview}.
% We introduce the preliminaries and pre-processing of an existing CSKG, ATOMIC$_{20}^{20}$~\cite{DBLP:conf/aaai/HwangBBDSBC21}, in \Cref{sec:preprocessing}. 
% We then introduce the definition and sampling of complex queries in \Cref{sec:query_sampling}. 

\subsection{Pre-processing}\label{sec:preprocessing}

We use ATOMIC$_{20}^{20}$~\cite{DBLP:conf/aaai/HwangBBDSBC21}, a comprehensive Commonsense Knowledge Graph covering everyday social, physical, and
event-level knowledge, as the base CSKG.
Before sampling queries, we address the sparsity and quality issues first.

\paragraph{Sparsity} 
CSKGs are usually highly sparse compared to factual KGs due to the diversity and scale of commonsense~\cite{DBLP:conf/aaai/MalaviyaBBC20}, 
% making it hard to sample diverse complex queries.
resulting in many isolated nodes that can hardly be sampled as part of a complex query.
To alleviate this issue, we develop a set of rules and use sentence embedding similarity to merge nodes in the CSKG, leading to 22.4\% of nodes being merged and an average degree increase of 25.3\%. 
In the final query sampling process, the number of 2p paths increased from 7,382 to 405,492, and the number of 2i queries rose from 1.43M to 2.06M.

\begin{figure}[t]
    \centering
    \includegraphics[width=1\linewidth]{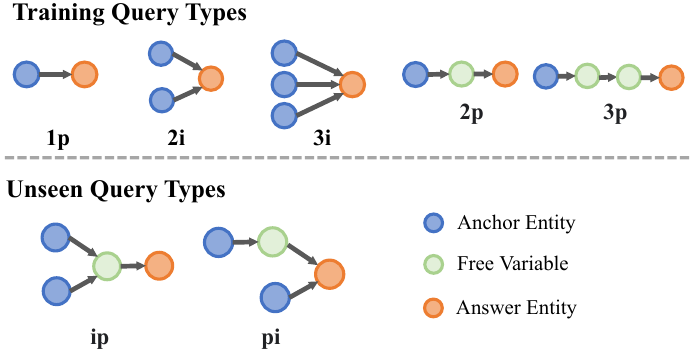}
    % \vspace{-0.15in}
    \caption{Visualization of query structures. The anchor entities and relations are specified to instantiate the query. `p' and `i' represent \textit{projection} and \textit{intersection}, and the number ahead of p and i indicates the number of anchor entities and free variables. }
    \label{fig:query_types}
    % \vspace{-1.2em}
\end{figure}

\begin{figure*}[t]
    \centering
    \includegraphics[width=1\linewidth]{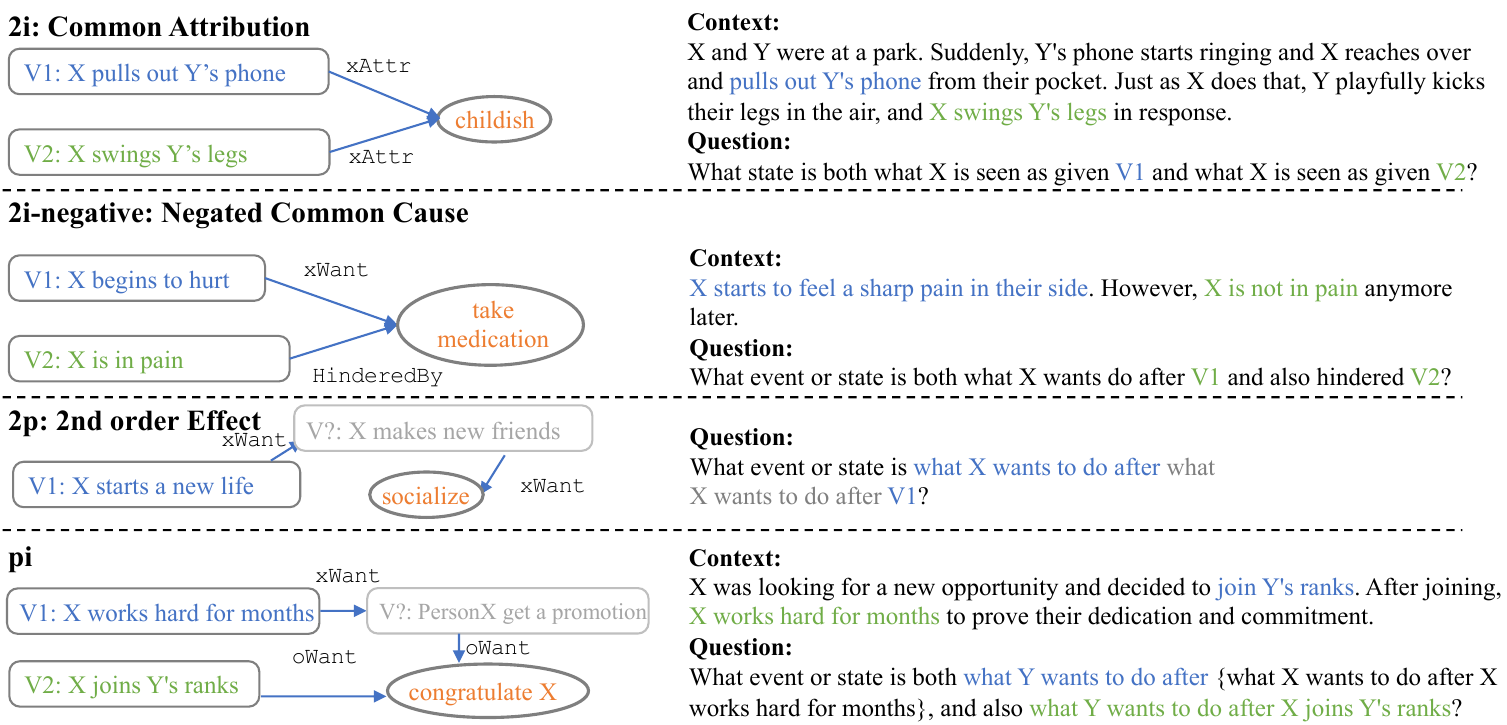}
    % \vspace{-0.1in}
    \caption{Examples of different query types, their verbalization, and corresponding questions. }
    \label{fig:examples}
    % \vspace{-0.2in}
\end{figure*}

\paragraph{Quality} 
The error rate of CSKGs (e.g., ATOMIC has an error rate of $\sim$10\%) can be problematic when we consider the intersection and projection of more than two triples (errors in a single triple could propagate to many multi-hop queries).
We use an off-the-shelf plausibility scorer Vera~\cite{Vera}, a T5-based scorer fine-tuned on 2 CSKGs and 19 QA datasets, to score every triple in terms of commonsense plausibility (between 0 to 1).
We filter out triples ($\sim$10\%) with a plausibility score lower than 0.5, the threshold provided in \citet{Vera} for plausible statements.

\subsection{Query Sampling}\label{sec:query_sampling}
% In this subsection, we introduce the complex queries that we study and the sampling of complex queries.
The query structures that we study are visualized in \Cref{fig:query_types}.
Following \citet{query2box}, we use projections (1p, 2p) and intersections (2i, 3i) as training queries, and leave complex queries ip and pi as zero-shot evaluation queries. 
To examine scenarios involving negation and differentiate them from regular 2i queries, we use the term ``2i-neg'' to represent 2i queries where one of the relations is ``HinderedBy''.
% In practice, due to the sparsity of ATOMIC$_{20}^{20}$ and the ambiguity of commonsense, we don't consider 3p as only a small number of 3p queries can be sampled and the causality becomes too weak to draw meaningful inferences when the reasoning chain becomes three-hop.
% Here, multi-hop projection corresponds to inferring hidden reasoning context, and the intersection operations requires reasoning about complex interactions between events. 
In this formulation, multi-hop projection involves inferring hidden reasoning contexts, while intersection operations require reasoning about complex interactions between events.

Given a query structure, we use pre-order traversal to sample free variables and anchor entities starting from an answer entity. 
We sample predecessors uniformly based on (relation, entity) pairs.
During sampling, to avoid over-sampling on nodes with extremely high degree, we empirically set a cut-off degree $\mathcal{T}=10$ to only sample from top $\mathcal{T}$ neighbors of a node scored by Vera.
In the end, we conduct a post-order traversal starting from the anchor entities to find all the answers of the query, in addition to the starting answer entity.
% More details are presented in \Cref{sec:appx_query_sampling}.

\paragraph{Distractor Sampling}
We sample 4 additional candidate distractors for each query, where 2 of them are randomly sampled across the whole CSKG, and 2 of them are sampled from the neighbors of the anchor entities that are not the answers to the whole query, represented as adversarial negative examples.
When fine-tuning a question answering model, 
the negative examples are used as synthetic question answering pairs for training.
In the evaluation set, these candidate negative examples, together with the sampled answer, are manually annotated to form a gold evaluation set.

\subsection{Verbalization}\label{sec:verbalization}

CSKGs are constructed in a context-free manner.
To make the logical queries on such context-free triples more human-interpretable, we introduce an additional step of verbalizing the anchor entities to a narrative, to effectively acquire fluent and plausible narrative-inference pairs.
\paragraph{Anchor Entity Verbalization}
We consider a rule-based verbalizer and a ChatGPT-driven verbalizer.
In the rule-based verbalizer, we add a discourse marker between the two or three anchor entities depending on the semantics of the query relations. 
For example, a simple situation would be adding an ``and'' or ``then'' between two anchor entities in a 2i query.
To make the query more human-understandable, we consider using ChatGPT to synthesize necessary contexts to make the query an actual narrative.
We include the detailed rules for adding discourse connectives, and prompts for using ChatGPT to verbalize complex queries in \Cref{sec:appx_verbalization}.

\begin{table*}[t]
\small
\centering
\begin{tabular}{l|cccccc|c}
\toprule
\textbf{Method} & \textbf{2i} & \textbf{2i-neg} & \textbf{3i} & \textbf{2p} & \textbf{ip} & \textbf{pi} & \textbf{All}\\ % & Avg. \\ 
\midrule
\multicolumn{1}{@{}l}{\textbf{API-based LLMs}}\\
\multicolumn{1}{@{}l|}{\texttt{gpt-3.5-turbo-0613}} & 33.56& 43.12& 42.01& 38.66& 38.05& 28.40& 37.74 \\
% - zero-shot 
- 1-shot & 43.31 & 35.31 & 58.45 & 57.73 & 51.33 & 62.96 & 48.22 \\
- 1-shot w/ CoT & 45.80 & 36.43 & 54.34 & 57.73 & 50.44 & 66.67 & 48.75  \\
% - 1-shot (2p) & 40.59 & 31.23 & 49.77 & 41.24 & 42.48 & 59.26 & 41.61 \\
% - 1-shot (2p) w/ CoT & 45.12 & 34.20 & 49.77 & 44.85 & 43.36 & 61.72 & 44.50  \\
% - 6-shot (2i) & 46.26 & 43.87 & 59.82 & 52.58 & 50.44 & 72.84 & 50.95 \\
% - 6-shot (2i) w/ CoT & 53.29 & 44.98 & 59.36 & 55.15 & 53.10 & 66.67 & 53.68  \\
- 8-shot (2i, 2p) & 48.52 & 41.26 & 57.08 & 67.53 & 53.10 & 74.07 & 53.22  \\
- 8-shot (2i, 2p) w/ CoT & 52.61 & 46.10 & 60.27 & 59.79 & 52.21 & 65.43 & 54.37 \\
% \midrule
\multicolumn{1}{@{}l|}{\texttt{gpt-4-1106-preview}} & 44.67 & 46.47 & 52.05 & 32.47 & 40.71 & 53.08 & 44.64 \\
% - zero-shot & 
- 1-shot & 47.85 & 42.01 & 50.68 & 38.66 & 44.25 & 50.62 & 45.63 \\
- 1-shot w/ CoT & 48.97 & 46.46 & 52.96 & 49.48 & 52.21 & 58.02 & 50.04 \\
- 8-shot (2i, 2p) & 54.87 & 46.47 & 58.90 & 45.88 & 52.21 & 66.67 & 53.00 \\
- 8-shot (2i, 2p) w/ CoT & 57.82 & 49.07 & 62.56 & 61.34 & 52.21 & 66.67 & 57.40 \\
\midrule
\multicolumn{3}{@{}l}{\textbf{Open-source (QA) Language Models}}\\
HyKAS (\citealp{DBLP:conf/aaai/MaIFBNO21}, zero-shot) & 34.92 & 39.41 & 27.85 & 41.75 & 37.17 & 33.33 & 35.76 \\
% CAR (zero-shot) & 35.37 & 34.94 & 34.70 & 47.94 & 30.97 & 38.27 & 36.83\\
CAR (\citealp{wang-etal-2023-car}, zero-shot) & 37.41 & 30.48 & 37.44 & 57.73 & 32.74 & 53.09 & 39.56 \\
Llama2 (7B)~\cite{DBLP:journals/corr/abs-2307-09288} & 35.15 & 21.93 & 39.27 & 35.57 & 28.32 & 51.85 & 33.64 \\
Vera (5B)~\cite{Vera} & 47.62 & 27.51 & 40.18 & 66.49 & 52.21 & 58.02 & 46.09 \\
UnifiedQA-v2~\cite{DBLP:journals/corr/abs-2202-12359} & 56.23 & 39.41 & 62.56 & 58.76 & 51.33 & 62.96 & 54.21 \\
Flan-T5 (11B)~\cite{DBLP:journals/corr/abs-2210-11416} & 58.28 & 47.21 & 65.30 & \textbf{76.29} & 56.64 & 79.01 & 60.97 \\
\midrule
\multicolumn{3}{@{}l}{\textbf{Fine-tuned on \dataname{}}}\\
DeBERTa-v3-Large (+\dataname{}) & 60.09 & \textbf{58.36} & 69.41 & 61.86 & \textbf{59.29} & 81.48 & 62.79 \\
CAR-DeBERTa-v3-Large (+\dataname{}) & \textbf{61.22} & 56.13 & \textbf{69.86} & 68.56 & 56.64 & \textbf{85.19} & \textbf{63.78} \\
\bottomrule
\end{tabular}
% \vspace{-0.1in}
\caption{Model performance (\%) on the multiple-choice question answering evaluation set of \dataname{}.}
\label{tab:mcqa_main_results}
% \vspace{-1em}
\end{table*}

\paragraph{Relation Verbalization} The multiple relations in complex queries can be deterministically converted to a question using the natural language descriptions of the relations, presented in \Cref{sec:appx_verbalization}.

\begin{table*}[t]
\small
\centering
\begin{tabular}{@{}l|c|lllll|l|@{}c@{}} %
\toprule
 \multirow{2}{*}{\textbf{Model}} & \multirow{2}{*}{\textbf{CSKG}} & \multicolumn{6}{c|}{\textbf{Out-of-domain}} & \textbf{\ In-dom.} \\
 % \cline{3-9}
 \cmidrule(lr){3-9}
 &  &  a-NLI & CSQA & PIQA & SIQA & WG & Avg. & \dataname{} \\ 
\midrule
Random & - & 50.0 & 20.0 & 50.0 & 33.3 & 50.0 & 40.7 & 20.0 \\
% Majority & - & 50.8 & 20.9 & 50.5 & 33.6 & 50.4 & 41.2 & 21.2 \\
% GPT2-L~\cite{radford2019language} & - & 56.5 & 41.4 & 68.9 & 44.6 & 53.2 & 52.9 \\
% RoBERTa-L~\cite{DBLP:journals/corr/abs-1907-11692} & - & 65.5 & 45.0 & 67.6 & 47.3 & 57.5 & 56.6 & 13.4 \\
DeBERTa-v3-L~\cite{he2023debertav} & - & 59.9 & 25.4 & 44.8 & 47.8 & 50.3 & 45.6 & 14.7 \\
Self-talk~\cite{DBLP:conf/emnlp/ShwartzWBBC20} & - & - & 32.4 & 70.2 & 46.2 & 54.7 & - & -\\
\textsc{Comet}-DynaGen~\cite{DBLP:conf/aaai/BosselutBC21} & ATOMIC & - & - & - & 50.1 & - & - & -\\
SMLM~\cite{DBLP:conf/emnlp/BanerjeeB20} & * & 65.3 & 38.8 & - & 48.5 & - & - & -\\
MICO~\cite{DBLP:conf/emnlp/SuWFZSZ22} & ATOMIC & - & 44.2 & - & 56.0 & - & - & -\\
STL-Adapter~\cite{DBLP:conf/naacl/KimKKAHY22} & ATOMIC & 71.3 & 66.5 & 71.1 & 64.4 & 60.3 & 66.7 & - \\
\midrule
\multicolumn{8}{@{}l}{\textbf{Large Language Models}} \\
GPT-3.5 (\texttt{text-davinci-003}) & - & 61.8 & 68.9 & 67.8 & {68.0} & 60.7 & 65.4 & - \\
GPT4 (\texttt{gpt-4-1106-preview}) & - & 75.0 & 43.0 & 73.0 & 57.0 & 77.0 & 65.0 & 44.6 \\
ChatGPT (\texttt{gpt-3.5-turbo}) & - & 69.3 & \underline{74.5} & 75.1 & \underline{69.5} & 62.8 & 70.2 & 37.7 \\
\ \ + zero-shot CoT & - & 70.5 & \textbf{75.5} & \underline{79.2} & \textbf{70.7} & 63.6 & 71.9 & 28.9\\
% \quad + Self-consistent chain-of-thought & - & 73.2 & \textbf{75.7} & \underline{81.7} & \underline{69.7} & 64.1 & 72.9 \\
% \ \ + eight-shot synthetic exemplar & ATOMIC & \\
\midrule
\multicolumn{8}{@{}l}{\textbf{Backbone: DeBERTa-v3-Large} \scriptsize{\textit{435M}}} \\
HyKAS~\cite{DBLP:conf/aaai/MaIFBNO21} & ATM-10X & 75.1 & 71.6 & 79.0 & 59.7 & 71.7 & 71.4 & 27.7 \\
HyKAS~\cite{DBLP:conf/aaai/MaIFBNO21} & ATOMIC & 76.0 & 67.0 & 78.0 & 62.1 & 76.0 & 71.8  & 35.8\\
% \hline
CAR \cite{wang-etal-2023-car} & ATOMIC & 78.9 & 67.2 & 78.6 & 63.8 & 78.1 & 73.3 & 36.8\\
CAR \cite{wang-etal-2023-car} & ATM$^{C}$ & 79.6 & 69.3 & 78.6 & 64.0 & \underline{78.2} & 73.9 & 39.8 \\
% \midrule
% \multicolumn{8}{@{}l}{\textbf{Ours: DeBERTa-v3-Large} \scriptsize{\textit{435M}}} \\
HyKAS + \dataname (Ours) & ATM, \dataname & 78.4 & 69.9 & 78.7 & 64.1 & \textbf{78.3} & \underline{73.9} & \underline{62.8} \\
CAR + \dataname (Ours) & ATM$^C_{\text{,}}$ \dataname & \textbf{81.2} &70.9 & \textbf{80.3} & 65.6 & 77.4 & \textbf{75.1}  & \textbf{63.8} \\
\midrule
% \midrule
% \multicolumn{8}{@{}l}{\textbf{Supervised Learning \& Human Performance}} \\
% RoBERTa-L (Supervised) & - & 85.6 & 78.5 & 79.2 & 76.6 & 79.3 & 79.8 \\
% DeBERTa-v3-L (Supervised) & - & 89.0 & 82.1 & 84.5 & 80.1 & 84.1 & 84.0 \\
Human Performance & - & 91.4 & 88.9 & 94.9 & 86.9 & 94.1 & 91.2 & - \\
\bottomrule
\end{tabular}
% \vspace{-0.1in}
\caption{Zero-shot evaluation results (\%) on five out-of-domain commonsense question answering benchmarks, and the in-domain evaluation set of \dataname{}. 
The best results are \textbf{bold-faced}, and the second-best ones are \underline{underlined}.
}
\label{tab:zeroshot_mcqa}
% \vspace{-0.15in}
\end{table*}

% \vspace{-1em}
\subsection{Human Annotation}\label{sec:human_annotation}

To support reliable automatic evaluation, we formalize the problem of complex commonsense reasoning as a multi-choice question answering task, with one true answer, three distractors, and a fifth option indicating ``None of the answers are correct''.
We crowdsourced the answers using Amazon Mechanical Turk (AMT). 
The workers are given the verbalized query as the context, the verbalized relations as the question, and the sampled (negative) answers.
% The workers are asked to determine whether each of the five options are correct or not.
If no sampled answers are correct, then the worker is asked to select an additional ``None of the answers are correct'' option.
If the verbalization itself does not make sense, the worker can also select another option ``The context doesn't make sense or is meaningless'' and we discard the example.
Each question is annotated by three workers. 
The workers are paid on average 16 USD per hour.
Our final dataset consists of $\sim$782k training examples and 1317 manually-validated evaluation examples. 

% \finalcopy{
\paragraph{Quality} The overall per-option inter-annotator agreement is 78\%, and the Fleiss kappa is 0.445, indicating moderate agreement. 
Among 1.3K verified examples, 4.7\% were labeled as incorrect contextualization. 
The likelihood that a sampled answer is the correct response to the contextualized question is 52.1\%. 
For randomly sampled negative examples and one-hop neighbors, the plausibility rate is 23.5\%, notably lower than the sampled answers. 
The authors of this paper manually checked the examples where the IAA between three annotators is lower than 0.6 and fixed the answers to ensure quality. 
A similar distribution is expected for the training set. Another thing to note that even though the training set is silver-standard, language models fine-tuned on it can autonomously identify patterns and acquire valuable insights from a large number of complex queries, resulting in improved reasoning performance, which will be shown in the next section.
% }

% While the training set is silver-standard and the probability that a sampled answer is plausible is not 100\%, language models can autonomously identify patterns and acquire valuable insights from a large number of distantly-labeled complex queries.  This effectiveness is evidenced by notable improvements in both generative commonsense inference and commonsense question answering across eight diverse datasets spanning various domains. 
% The evaluation set is gold-standard after human annotation and is reliable for evaluation.

%
More details can be found in \Cref{sec:appx_data_construction}.

\section{Experiments}\label{sec:exp}

We conduct experiments on the evaluation set of \dataname{}, formulated as a Multi-Choice Question Answering (MCQA) task. 
Specifically, we examine the performance of state-of-the-art off-the-shelf language models on \dataname{}, and also study the effect of training a question answering model on the distantly supervised training set of \dataname{}.

\subsection{Setup}

We use popular API-based and open-source LLMs as baselines.
Following the standard practice of prompting LLMs for QA~\cite{DBLP:journals/corr/abs-2210-12353}, 
% we ask LLMs to output the associated symbol (e.g, `A' and `B')in from the options.
we initialize a prompt that takes ``[Context] [Question] [Options]'' as the input and ask the model to only output the associated symbol (e.g., `A') in the QA pair as the prediction.
For open-source language models like Flan-T5 and Llama2, we use the same prompt, and compute the logits received by each of the options in the first prediction token.

We also study the effect of fine-tuning a question-answering model on the synthetic training queries discussed in \Cref{sec:query_sampling}. 
We follow the pipeline by HyKAS~\cite{DBLP:conf/aaai/MaIFBNO21}, 
which fine-tunes language models on QA pairs synthesized from one-hop knowledge in CSKGs, and extend it to complex queries.
For one-hop (1p) triples, the head and relation are transformed into a question with pre-defined prompts.
% For example, the triple (\textit{PersonX played a football game}, \texttt{xWant}, \textit{take a rest}) in ATOMIC can derive a question ``\textit{What does PersonX want after played a football game?}'' with the correct answer as ``\textit{take a rest}''.
For complex queries, the verbalized queries (as illustrated in \Cref{sec:verbalization}) are regarded as the context, and questions are also transformed with a different prompt template depending on the relations.
The tails to the one-hop triple or the sampled answer to the query are regarded as the correct answer, and the negative examples are randomly sampled across the whole CSKG following a keyword overlapping filtering~ \cite{DBLP:conf/aaai/MaIFBNO21, wang-etal-2023-car}.
We use DeBERTa-v3-large as the backbone encoder.\footnote{We refer readers to \Cref{sec:appx_exp} for detailed implementations and prompt templates.}
% , and train the QA model for 1 epoch with a learning rate of 5e-6 and a linear learning rate decay.

\begin{table*}[t]
\small
\centering
\setlength\tabcolsep{3pt}
\begin{tabular}{@{}l|c|cc@{}c@{}|cc@{}c@{}||cc@{}c@{}||cc@{}c@{}}
\toprule
\multirow{2}{*}{\textbf{Model}} & \multirow{2}{*}{\textbf{Training Data}}  & \multicolumn{3}{c|}{\textbf{Multi-Event}} &  \multicolumn{3}{c||}{\textbf{Paragraph-Level}} & \multicolumn{3}{c||}{\textbf{Single-Event}} & \multicolumn{3}{c}{\textbf{\dataname{}}}\\
\cmidrule(lr){3-14}
 &  & \footnotesize{\textbf{B-2}} & \footnotesize{\textbf{R-L}} & \footnotesize{\ \textbf{BERT}} & \footnotesize{\textbf{R-L}} & \footnotesize{\textbf{CIDE}} & \footnotesize{\ \textbf{BERT}} &  \footnotesize{\textbf{R-L}} & \footnotesize{\textbf{CIDE}} & \footnotesize{\ \textbf{BERT}} & \footnotesize{\textbf{R-L}} & \footnotesize{\textbf{CIDE}} & \footnotesize{\ \textbf{BERT}} \\ % & Avg. \\ 
\midrule
\multicolumn{2}{l}{\textbf{(Distantly) Supervised Learning}} \\
\midrule
COMET-M (BART-L) & MEI &   25.1 & 33.6 & 64.9 & - & - & - & - & - & - & - & - & - \\
COMET-M (GPT-2-L) & MEI  & 16.2 & 25.7 & 55.1 & - & - & - & - & -& - & - & - & -\\
ParaCOMET (GPT-2-L) & PCD  & - & - & - & 18.8 & 27.8 & 60.2 & - & - & - & - & - & -\\
\midrule
\midrule
\multicolumn{8}{l||}{\textbf{Zero-shot Learning}}&\multicolumn{3}{l}{\textbf{Supervised}} \\
\midrule
% \textbf{Zero-shot} \\
COMET &1p  & 1.20  & 2.73 & 38.9 & 3.5 & 6.4 & 25.7 & 50.0 & 66.1 & 75.1 & 10.0 & 20.7 & 44.3 \\
COMET-distill &ATM10x &  1.20 & 3.55 & 12.7 & 11.8 & 16.8 & 29.5 & 1.6 & 4.8  & 24.3 & 8.3 & 11.9 & 36.1 \\
% COMET-distill &1p, ATM10x &  \\
\dataname-COMET & 1p, 2i  & \textbf{8.87} & \textbf{15.2} & \textbf{46.4} & \textbf{13.8} & \textbf{22.1} & \textbf{53.7} & \textbf{50.7} & \textbf{68.0} & \textbf{77.1} & 13.6 & 26.1 & 39.8 \\
\dataname-COMET & 1p, 2p, 2i, 3i & 5.41  & 10.4 & 44.8 & 9.2 & 16.6 & 44.1 & 50.4 & 66.9 & 77.1 & \textbf{14.7} & \textbf{33.0} & \textbf{46.3} \\
\midrule
LLama2-7b & -  & 1.81 & 4.14 & 45.7 & 2.2 & 2.2 & 48.6 & 5.4 & 2.9 & 51.5 & 3.9 & 6.7 & 44.9 \\
COMET-LLama2-7b & 1p  &7.62  & 14.4 & 44.2 & 9.1 & 12.3 & 51.0 & 27.5 & 26.4 & 64.2 & 10.9  & 22.3 & 44.9 \\
\dataname-LLama2-7b & 1p, 2i  &\textbf{8.82}  & \textbf{16.4} & \textbf{47.5}  & 14.6 & \textbf{22.1} & 55.3 & \textbf{31.6} & \textbf{31.1} & \textbf{66.0} & \textbf{35.7} & \textbf{107.2} & \textbf{61.3} \\
%ChatGPT: Bleu-2: 7.529   Bleu-4: 3.371   Rouge-L: 14.436 BertScore: 0.464
% chat paracomet: 14.8 & 23.6 & 55.5
\dataname-LLama2-7b & 1p, 2p, 2i, 3i  & 8.22  & 15.4 & 47.0 & \textbf{15.9} & 21.3 & \textbf{55.3} & 31.3 & 29.8 & 65.5 & 35.6 & 105.0 & 60.1 \\
% \midrule
% Zero-Shot &  0 & 0 & 0 & 0 & 0.1 & 0 & 0 & 0.0 \\
% \textbf{Supervised Learning} \\
\bottomrule
\end{tabular}
% \vspace{-0.1in}
\caption{Experimental results on downstream narrative commonsense reasoning, including in a multi-event~\cite{comet-m} setting, and a paragraph-level setting~\cite{paracomet}. In-domain settings include single-event generation and complex inference in \dataname{}. We use BLEU-2 (B-2), ROUGE-L (R-L), CIDEr (CIDE), and BERTScore (BERT) as the evaluation metrics.}
\label{tab:comet_performance}
% \vspace{-0.1in}
\end{table*}

\subsection{Results and Analysis}

Our results are presented in \Cref{tab:mcqa_main_results}.
% In terms of performance on commercial LLMs, GPT-4 generally outperforms ChatGPT by a notable margin.
We observe that Chain-of-Thought (CoT) improves reasoning performance, as it allows the model to first induce the causes or effects of individual events in intersection-based queries (2i and 3i), or induce hidden variables in projection-based queries (2p as in \Cref{fig:query_types}).
% The eight-shot CoT, which encompasses both 2i and 2p queries as exemplars, yields the highest performance due to the coverage of all base query types. 
Adding eight-shot exemplars (consisting of 2i, 2i-neg, and 2p queries) further improves performance among prompting baselines.

% By pre-training on the synthetic QA pairs of 1p, 2i, 3i, and 2p queries, we can acquire a stronger QA model based on the question-answer pipeline~\cite{DBLP:conf/aaai/MaIFBNO21}.
For models fine-tuned on complex queries using HyKAS and CAR, we observe that the synthetic training pairs, despite lacking manual annotation, serve as valuable distant supervision signals. 
They enhance the complex reasoning capability of HyKAS and CAR, surpassing the performance of the 8-shot GPT-4 model with CoT by 6\%.
CAR + \dataname{} also outperforms the 11B version of UnifiedQA-v2 and Flan-T5, which are both fine-tuned on numerous (commonsense) question answering datasets, by 9\% and 3\%, respectively.
% In terms of fine-tuning on complex queries based on the paradigm of HyKAS and CAR, we can see that even though the synthetic training pairs are not further manually annotated, they are useful distant supervision signals to boost the complex reasoning ability of a QA model, even achieving 6\% better than 8-shot GPT-4 with CoT. 
% We also include the zero-shot transferability experiments of this QA model to other commonsense QA datasets, which will be presented in \Cref{sec:exp_zs_csqa}.

\section{Downstream Evaluation}

In addition to benchmarking Complex Commonsense Reasoning, we also study the effect of leveraging \dataname{} as training data to generalize to other downstream commonsense reasoning tasks.
As tasks, we use zero-shot CommonSense Question Answering (CSQA), and Generative Commonsense Inference, including one-hop, multi-event, and paragraph-level settings.

\subsection{Commonsense Question Answering}\label{sec:exp_zs_csqa}

\paragraph{Setup}
The task of zero-shot commonsense QA involves selecting the most plausible option for commonsense questions without training on examples from the benchmark dataset.
We directly leverage the model we trained in \Cref{sec:exp}, 
the DeBERTa-v3-large-based model fine-tuned on synthetic question pairs from both ATOMIC and \dataname{}, and check the performance on five popular commonsense question answering datasets: Abductive NLI (aNLI;~\citealp{DBLP:conf/iclr/BhagavatulaBMSH20}), CommonsenseQA (CSQA;~\citealp{DBLP:conf/naacl/TalmorHLB19}), PhysicalIQA (PIQA;~\citealp{DBLP:conf/aaai/BiskZLGC20}), SocialIQA (SIQA;~\citealp{DBLP:conf/emnlp/SapRCBC19}), and WinoGrande (WG;~\citealp{DBLP:journals/cacm/SakaguchiBBC21}).
As baselines, we consider the same methods, HyKAS \citep{DBLP:conf/aaai/MaIFBNO21} and CAR \citep{wang-etal-2023-car}, but use other CSKGs as training sets. 
In \Cref{tab:zeroshot_mcqa}, ATM-10X refers to ATOMIC-10x from \citet{DBLP:conf/naacl/WestBHHJBLWC22}, and ATM$^C$ refers to the training data from CAR~\cite{wang-etal-2023-car} which is augmented from ATOMIC with conceptualization.

\paragraph{Results and Analysis}

We report model performance in \Cref{tab:zeroshot_mcqa}. 
We observe the inclusion of \dataname{} and one-hop triples from ATOMIC as training data for CAR and HyKAS yields significant improvements in question answering ability. 
% This improvement is observed in both in-domain complex reasoning tasks and out-of-domain CSQA tasks. 
Notably, the combination of CAR and \dataname{} achieves the highest performance among all models, surpassing even ChatGPT and GPT-4, despite having a parameter size at least two orders of magnitude smaller.

Notably, when using CAR as the base model, training on \dataname{} leads to the highest performance gain of around 1.8\% for a-NLI.
When evaluating on a-NLI, which includes instances of abductive reasoning, the model may be helped by learning from 2i queries where one relation represents \textit{cause} and the other represents \textit{effect} (abduction examples in \Cref{fig:Introduction_demo} and \Cref{fig:examples}).
% Similarly, in PIQA, which studies the relation between an action and a goal, complex event queries may provide clear be also reflects event-event relations where complex queries can provide a lot.
% For CSQA and SIQA, the performance is improved because the CSKG that CSQA and SIQA rely on is a subset of ATOMIC$_{20}^{20}$, on which we build our \dataname{}. 
Meanwhile, the performance on WinoGrande was adversely affected, likely because Winogrande primarily focuses on identifying distinguishing features of entity pairs. The benefits from learning event-event interactions from \dataname{} may not transfer well to this setting.

\subsection{Generative Commonsense Inference}\label{sec:exp_comet}

\paragraph{Setup} We study generative commonsense inference as an additional evaluation task. 
We include multi-event commonsense generation (COMET-M; \citealp{comet-m}) and paragraph-level commonsense generation (ParaCOMET; \citealp{paracomet}) as two out-of-domain evaluation tasks.
% COMET-M requires generating the commonsense causes of effects based on one of the several events in the context.
% ParaCOMET requires generating commonsense inferences of one sentence based on narrative from ROCstory, which typically contains four to five sentences.
We also include the vanilla COMET~\cite{DBLP:conf/acl/BosselutRSMCC19} as an additional in-domain evaluation, which focuses on 1p queries that require generating the tail given head and relation as the input.
% We report the generation performance on generative \dataname{} in the last columns.
We also conduct experiments on the generative sub-task of \dataname{}, where verbalized context and question inputs are used to inferences. 
The annotated ground answer options are used as references.

For the (distantly) supervised learning baselines, we fine-tune GPT-2-large on the annotated multi-event inference dataset (MEI) from \citet{comet-m} and distantly labeled PCD dataset from \citet{paracomet} as a reference.
In our zero-shot learning setting,
we study the effect of fine-tuning COMET (GPT-2-large) on ATOMIC and different query types of \dataname{}.
We also study fine-tuning an LLM, Llama2-7b, by converting triples and queries to an instruction-tuning format, following the prompt template in \Cref{sec:verbalization} and \Cref{sec:appx_generative_experiments}.
We leverage the framework of \citet{DBLP:journals/corr/abs-2311-16079}\footnote{https://github.com/epfLLM} to fine-tune Llama2-7b.
We fine-tune on a mixture of different query types as detailed in the \textbf{Training Data} column. 
We present the performance results of models fine-tuned on either the annotated or distantly supervised training set for both tasks as reference benchmarks. 
Specifically, we use MEI for COMET-M and PCD for ParaCOMET.
To ensure diversity and prevent overfitting to common tails, complex queries are selected using an n-gram based diversity filter~\cite{DBLP:conf/emnlp/YangMFSBWBCD20}.

\paragraph{Results and Analysis}
We present the results in \Cref{tab:comet_performance}. 
Compared to models fine-tuned solely on one-hop triples, COMET models fine-tuned on additional complex queries demonstrate enhanced generative commonsense inference capabilities for multi-event and paragraph-level scenarios. 
% This is because verbalized complex queries offer longer and more complicated reasoning contexts. These contexts align with the multi-event or paragraph-level inferences that involve multiple events and complex contextual information.
% The first two settings are out-of-domain complex commonsense reasoning tasks that require reasoning on longer context and more complicated event-event relations.
When comparing different query types, fine-tuning solely on 2i queries yields the most significant improvement in reasoning capability, likely because 2i queries provide more explicit reasoning signals compared to 2p queries, which can be ambiguous due to the large candidate space of the hidden event. 
For example, the average number of answers for 2p queries is 7.93, compared with 1.09 for 2i queries.
In addition, the answers to 2i queries exhibit greater diversity than 3i queries, as the CSKG is sparse and provides a limited number of distinct tails for sampling 3i queries compared to 2i queries.
% In addition, among different query types, 2i is the most useful query type that help improve the reasoning ability.
% This observation can be attributed to the nature of tasks in COMET-M and ParaCOMET, which do not heavily rely on second-order projection inference. 
% Instead, they mainly require the reasoning ability brought by intersection-based queries.

\begin{table}[t]
\centering
\small
\renewcommand\arraystretch{1.0}
\newcolumntype{C}[1]{>{\centering\arraybackslash}p{#1}}
\begin{tabular}{l|cccc}
\toprule
\multirow{2}{*}{\textbf{Model}} & \multicolumn{3}{c}{\dataname }  \\
& \footnotesize{\textbf{R-L}} & \footnotesize{\textbf{CIDEr}} & \footnotesize{\textbf{BERT}}\\
\midrule
\multicolumn{2}{l}{\textbf{Filter}} \\
\dataname{}-COMET & 14.7 & 33.0 & 46.3 \\
\ - w/o plau. filter & 13.0 & 31.2 & 42.3 \\
\ - w/o div. filter & 14.4 & 32.5 & 45.8 \\
\ - w/o both filter & 12.5 & 30.3 & 40.1 \\
\midrule
\multicolumn{2}{l}{\textbf{Query Types}} \\
COMET (1p) & 10.0 & 20.7 & 44.3 \\
\ + 2i & 13.6 & 26.1 & 39.8 \\
\ + 2p & 9.8 & 19.9 & 43.4 \\
\ + 2i, 3i, 2p & 14.7 & 33.0 & 46.3 \\
\midrule
\multicolumn{2}{l}{\textbf{Verbalization}} \\
\dataname{}-COMET & 13.6 & 26.1 & 39.8 \\
\dataname{}-COMET (V) & 14.3 & 27.1 & 43.4 \\
\dataname{}-Llama & 35.7 & 107.2 & 61.3 \\
\dataname{}-Llama (V) & 36.2 & 105.4 & 61.4\\
\midrule
\midrule
\multirow{2}{*}{\textbf{Model}} & \multicolumn{3}{c}{PCD}  \\
& \footnotesize{\textbf{R-L}} & \footnotesize{\textbf{CIDEr}} & \footnotesize{\textbf{BERT}}\\
\midrule
\multicolumn{2}{l}{\textbf{Verbalization}} \\
\dataname{}-COMET & 13.8 & 22.1 & 53.7 \\
\dataname{}-COMET (V) & 14.0 & 23.2 & 54.0 \\
\dataname{}-Llama & 14.6 & 22.1 & 55.3 \\
\dataname{}-Llama (V) & 14.8 & 23.6 & 55.5\\
\bottomrule
\end{tabular}
% \vspace{-0.1in}
\caption{Ablation studies on filters, type of queries, and using ChatGPT for verbalizing queries (denoted as V).}
\label{tab:ablation}
% \vspace{-1em}
\end{table}

\section{Analysis \& Discussion}

\subsection{Ablation Study}

We analyze the impact of various data filters, query types, and verbalization methods on generative inference within \dataname{}. Detailed results can be found in \Cref{tab:ablation}. % in the appendix.

\paragraph{Filtering}
We include two types of filters, a Vera-based plausibility filter and a diversity filter.
Evaluating the performance of generative commonsense inferences on \dataname{}, we examine the impact of removing both filters while employing GPT2-Large as the backbone model. 
Removing the plausibility filter results in a significant performance decline, highlighting its critical role. 
On the other hand, the diversity filter exhibits a minor positive influence on enhancing performance.

\paragraph{Type of Queries}
We investigate the impact of training our models on different types of logical queries.
The model trained only on 1p and 2p queries does not generalize well to other query types such as pi and ip, leading to a worse performance than the model trained on all query types.
However, according to \Cref{tab:mcqa_main_results} and \Cref{tab:comet_performance}, models trained on only 2i queries generalize better to downstream commonsense reasoning tasks, potentially indicating that multi-event reasoning in most existing commonsense benchmarks focus on intersection more than projection.

% most existing commonsense benchmarks focusing on interactions regarding multiple events are actually structured as an intersection-based manner, instead of projections and more complicated structures.
% such as ip and pi.

\paragraph{Verbalization}
We investigate the effect of using a rule-based verbalizer or ChatGPT-enabled verbalizer to generate \dataname{} contexts. 
% as detailed in \Cref{sec:appx_verbalization}.
% We found that we can achieve similar results with both verbalizer, 
% indicating that the rule-based verbalization is enough for LLMs to unlock complex reasoning ability.
Using ChatGPT-verbalized queries leads to better downstream performance on both PCD and \dataname{}.
In \dataname{}, the presence of ChatGPT-verbalization intuitively improves performance since the training context aligns with the evaluation set's format. 
On the other hand, the context in the PCD dataset is long and comprised of five sentences. 
Verbalization not only adds more contexts to the training but also aligns better with the PCD format.

\begin{table}[t]
\centering
\small
\renewcommand\arraystretch{1.0}
\newcolumntype{C}[1]{>{\centering\arraybackslash}p{#1}}
\begin{tabular}{l|cccc}
\toprule
Model & \#Plau. & \#1-hop & \#False\\
\midrule
LLama2-7b &  26& 2 & 28  \\
COMET-LLama2-7b & 29 & 8 & 23 \\
\dataname-LLama2-7b (2i) & 47 & 2 & 11\\
\dataname-LLama2-7b (all) & 45 & 3 & 12\\
\bottomrule
\end{tabular}
% \vspace{-0.1in}
\caption{Human evaluation results on the generative sub-task in \dataname{} using Llama2-7b as the backbone. `1-hop' indicates the answer is plausible in terms of only one-hop relations.}
\label{tab:error_analysis}
% \vspace{-1em}
\end{table}

\subsection{Error Analysis}
% We present a human-annotated quality evaluation on Llama-7b-based model on generative \dataname{} in \Cref{tab:error_analysis}.
We present a human-annotated quality evaluation of the Llama-7b-based model on the generation sub-task of \dataname{}.
To ensure diverse coverage of query types, we randomly sampled 60 queries, with 10 from each of the 6 types.
Manual inspection revealed a common error where the generated output was partially correct, either providing the answer to one of the triples in an intersection query or only the one-hop answer instead of the two-hop answer in 2-projection (2p) queries.
\Cref{tab:error_analysis} includes the number of such `1-hop' partially correct answers. 
Our results demonstrate that the zero-shot Llama model already produces 26 out of 60 plausible inferences. 
Fine-tuning the model on one-hop ATOMIC further increases the number of plausible generations while more frequently generating inferences that are one-hop correct. Moreover, fine-tuning on the synthetic training set of \dataname{} significantly improves the model's ability to generate complex commonsense inferences and reduces the occurrence of partially correct answers.
We provide case studies in \Cref{sec:appx_error_analysis}.

\section{Conclusion}

In this paper, we leverage the concept of conjunctive logical queries to create a complex commonsense reasoning dataset derived from CSKGs.
The dataset, \dataname{}, comprises a human-annotated evaluation set and a distantly supervised training set without further annotations.
Our experiments highlight the challenging nature of complex commonsense reasoning that involves multiple events or multi-hop scenarios, even for advanced language models such as GPT-4.
Additionally, we train question answering models and generative commonsense reasoning models using \dataname{}. The results show significant improvements across eight diverse downstream commonsense reasoning tasks, %, encompassing various aspects. 
highlighting the potential of leveraging CSKGs to acquire complex reasoning signals inexpensively, without relying on extra human effort.

% \newpage

\section*{Acknowledgement}

Yangqiu Song was supported by the NSFC Fund (U20B2053) from the NSFC of China, the RIF (R6020-19 and R6021-20), and the GRF (16211520 and 16205322) from RGC of Hong Kong. 
Yangqiu Song thank the support from the UGC Research Matching Grants (RMGS20EG01-D, RMGS20CR11, RMGS20CR12, RMGS20EG19, RMGS20EG21, RMGS23CR05, RMGS23EG08). 
We thank the support from the Tencent AI Lab Rhino-Bird Focused Research Program.
We also gratefully acknowledge the support of the Swiss National Science Foundation (No. 215390), Innosuisse (PFFS-21-29), the EPFL Science Seed Fund, the EPFL Center for Imaging, Sony Group Corporation, and the Allen Institute for AI.

\section*{Limitations}

\paragraph{Data Construction}
The construction of \dataname{} relies on sampling complex logical queries from existing CSKGs, which requires addressing sparsity, quality, contextualization issues. Despite conducting normalization and filtering, there may still be missing links within ATOMIC and mislabeled or ambiguous triples, which limits the quality of our sampled queries.
Future works can focus on deriving complex queries from CSKGs with better quality and more diverse semantics, which should also have higher density, such as on ATOMIC-10x, NovATOMIC~\cite{west-etal-2023-novacomet}.

\paragraph{Evaluation} 
In the context of generative commonsense reasoning, we employ lexical-overlap based automatic evaluation metrics to assess the performance of the model in a scalable manner. However, since each query typically has 1 to 3 gold references on average, this type of evaluation may not accurately capture the true plausibility of commonsense inferences, which is inherently open-ended.
To address this limitation, we have supplemented the automatic evaluation with human annotation on a subset of sampled queries, but this approach is not scalable.

% Future research can focus on the development of automatic complex reasoning protocols based on large language models. Such protocols can delve into more fine-grained aspects such as typicality and the degree of correctness, even if it's only partially correct. 
% By incorporating these fine-grained considerations, we can enhance the evaluation of commonsense reasoning systems.

\section*{Ethical Considerations}

We sample the data from ATOMIC$_{20}^{20}$, which is an open-source commonsense knowledge graph that may contain biases around gender, occupation, and nationality~\cite{DBLP:conf/emnlp/MehrabiZMPRG21}. When constructing \dataname{}, these biases may propagate if biased triples are sampled in a complex query that becomes of the training set. 
% The dataset does not contain specific individuals or organizations. Instead, it employs generic placeholders such as PersonX, PersonY, and randomly replaced first names to represent subjects and objects.
% However, this paper primarily focuses on complex reasoning based on knowledge, which is in contrast to works that solely rely on one-hop biased knowledge exploitation. 
We collected 1.3k inferences through crowdsourcing. The participants were compensated with an hourly wage of 16 USD, which is comparable to the minimum wages in the US. The qualification was purely based on the workers' performance on the evaluation set, and we did not collect any personal information about the participants from MTurk.

\bibliography{anthology,custom}

\clearpage
\appendix

\section{Additional Details on Data Construction} \label{sec:appx_data_construction}

In this section, we provide additional details to node normalization, plausibility filter, verbalization, and human annotations.
The overview of our construction framework is presented in \Cref{fig:overview}.

\subsection{Nodes Normalization (Dealing with Sparsity)}
\label{sec:appx_node_normalization}

To alleviate the sparsity issue, we first normalize the tail entities with simple rules similar with that in Dense-ATOMIC~\cite{dense-atomic} and CKBP~\cite{DBLP:conf/emnlp/FangWCHZSH21}.
In ATOMIC, heads are pre-defined complete sentences (for example, ``PersonX says sorry'')
while tails are usually short phrases without a subject (for example, ``to say sorry'').
This discrepancy produces many duplicated nodes and make the graph sparser. 
We develop simple rules to add ``PersonX'' or ``PersonY'' in front of the tails to make them a complete sentence, if the tail does not have a subject. 
This process merged 3.7\% nodes together.
% We present the normalization rules in \Cref{table:normalization}. 
% Our goal and techniques align with the objective and the method of that in Dense-ATOMIC~\cite{dense-atomic} and CKBP~\cite{DBLP:conf/emnlp/FangWCHZSH21}, but is rather simplified. One notable difference is that we do not process tails that are already complete sentences. 
% These distinctions would bring only minor differences in the final node merging, as ``she goes home'' will be semantically similar to ``PersonX goes home'' when using MPNet embeddings to aggregate similar nodes later.

Second, as the nodes in ATOMIC are free-text, some nodes with the same semantic meaning are represented as separated nodes due to some minor annotation distinctions and errors, e.g., ``PersonX buys a ticket'' versus ``PersonX buys a ticket .''. 
These discrepencies can be addressed using embedding similarities~\cite{DBLP:conf/acl/WuSX23}.
We use a state-of-the-art sentence embedding model\footnote{https://huggingface.co/sentence-transformers/all-mpnet-base-v2}, to merge nodes with cosine similarity score over 0.95. 
In this process, 20.0\% nodes are merged together and the average degree increases by 25.3\%.

% For example, a tail of ``to go'' under the relation xWant will be transformed to ``PersonX go''.
% A tail of ``satisfied'' under the relation xAttr will be transformed to ``PersonX is satisfied''.

\begin{table}[h]
\small
\begin{tabular}{m{2.2cm}<{\centering} | m{4.5cm}}
  \toprule
  \multicolumn{1}{c|}{Relations} & Mapping rules \\
%   \multicolumn{1}{m}{}\\
  \hline 
  \tabincell{c}{xWant/oWant/\\xIntent/xNeed} & Add PersonX/Y in front of the tail and remove the initial ``to''  \\
  xEffect/oEffect & Add PersonX/Y in front of the tail \\
  xReact/oReact & Add PersonX/Y and ``is'' in front of the tail \\
  xAttr& Add a PersonX/Y and ``is'' in front of the tail \\
  \bottomrule
 \end{tabular}
 \caption{Normalization rules for ATOMIC tails.}
 \label{table:normalization}
  % \vspace{-5ex}
\end{table}

\subsection{Data Filtering}
\label{sec:appx_mislabel_filter}

\paragraph{Plausibility Filter} We verbalize a $(h, r, t)$ triple from ATOMIC using the default template as provided in \citet{DBLP:conf/aaai/HwangBBDSBC21}. 
For example, (PersonX repels PersonY's attack, xAttr, brave) would be transformed to a declarative statement ``If PersonX repels PersonY's attack, then PersonX is seen as brave''.
To obtain a plausibility score, we input the statement into the Vera-5B model.
0.5 is used as the threshold to draw a boundary between plausible and implausible statements. 
We perform a manual inspection on the triples scored by Vera and randomly select 40 samples for three plausibility score intervals. 
Among these, we find that 4/40 triples are plausible when the Vera scores range from 0 to 0.1. 13/40 triples are considered plausible within the score range of 0.2 to 0.25. 
Furthermore, we identify 20/40 triples as plausible when their plausibility scores hover around 0.5, when most of the triples are quite ambiguous.
By setting the filter threshold as 0.5, we filter out around 14\% triples that are of a relatively lower quality.

\paragraph{Diversity Filter}

To prevent overfitting to common tails, we conduct a diversity-based filter to acquire diverse queries for training.
We take inspirations from G-DAUG~\cite{DBLP:conf/emnlp/YangMFSBWBCD20}, to use a simple greedy algorithm to iteratively select training data, which has been proven useful for selecting augmented data.
To be more specific, for each unique answer, we adopt an iterative approach to select the verbalized query that contributes the highest number of unique 1-gram terms to an ongoing vocabulary constructed for each answer.
We select top-20 queries for each unique answer entity.

% \subsection{Query Sampling}
% \label{sec:appx_query_sampling}

\subsection{Verbalization}
\label{sec:appx_verbalization}

\paragraph{Query Verbalization}
We employ two methods to verbalize complex queries: a rule-based method and a ChatGPT-based method.

In the case of 2i and 3i queries, the rule-based method typically involves inserting an ``and'' between the anchor entities. 
However, if the query suggests a specific chronological order between the two events, we use ``then'' to connect the events.
For instance, in 2i queries where one triple is ($V_1$, xEffect, $V_?$) and the other is ($V_2$, xIntent, $V_?$), it implies that $V_?$ serves as the effect of $V_1$ and the intermediate hidden cause of $V_2$. In this scenario, $V_1$ should occur before $V_2$. 
Therefore, the verbalization would be ``$V_1$ then $V_2$''.

% We have two methods for verbalizing complex queries, a rule-based method and ChatGPT-based method. 
% For 2i and 3i, in most cases, we add an ``and'' between the anchor entities. 
% For the case where it suggests a specific chronological order of the two events, we add a ``then'' between the anchor entities.
% For example, in 2i queries, if one triple is ($V_1$, xEffect, $V_?$) and the other is ($V_2$, xIntent, $V_?$), it indicates $V_?$ is the intermediate hidden cause of $V_2$ and the effect of $V_1$. 
% In this case, $V_1$ should happen before $V_2$.
% As a result, the verbalization would be $V_1$ then $V_2$.

\begin{table*}[t]
\small
\centering
\newcolumntype{C}[1]{>{\centering\arraybackslash}p{#1}}
\begin{tabular}{m{7em}|m{40em}}
\toprule
Query & Prompt \\
\midrule
2i, ip, pi & \tabincell{p{40em}}{Given two events, come up with concise and necessary context to make the a coherent and understandable narrative. No more than 2 additional piece of context should be added. If the one of the given events itself is ambiguous and hardly make sense even with extra context, return NA. If the two events are totally irrelevant even with additional context, then simply return NA. If the given two events can be directly composed to a narrative with simple a discourse connective without additional context, then there's not need to add additional context.\slashn Mark the location of both events with <E1></E1> for event 1 and <E2></E2> for event 2 in the generated narrative. } \\
\midrule
2i-neg & Given two events, create a cohesive narrative by incorporating event 1 (E1) and negated event 2 (E2) to make the a coherent and understandable narrative. No more than 2 additional piece of context should be added. If the one of the given events itself is ambiguous and hardly make sense even with extra context, return NA. If the two events are totally irrelevant even with additional context, then simply return NA. If the given two events can be directly composed to a narrative with simple a discourse connective without additional context, then there's not need to add additional context.\slashn Mark the location of both events with <E1></E1> for event 1 and <E2></E2> for event 2 in the generated narrative.\slashn Don't explain the reasons why E2 didn't happen!!\slashn Remember that negating an event means stating that it did not occur. For instance, if event 2 is ``PersonX goes shopping,'' the negated form would be ``PersonX didn't go shopping''. \\
\bottomrule
\end{tabular}
\caption{System instructions for verbalizing complex queries given different query types.}
\label{tab:verbalization_prompt}
\end{table*}

For ChatGPT verbalization, we present the system instructions for verbalizing different kinds of queries in \Cref{tab:verbalization_prompt}. Then, we generate the verbalized contexts with six exemplars that are manually annotated.
In the system instruction, we also ask ChatGPT to output ``NA'' if the given anchor entities are totally irrelevant or too ambiguous.
We filter out those queries where the output is ``NA''.

For example, to better interpret the query in \Cref{fig:Introduction_demo}, we need to take into consideration both the relations of interest and the anchor entities.
The query asks about the effect of the first event and what causes (intention) of the second event, which is inherently represents \textit{abductive reasoning}. 
This requires the second event to happen before the first event, to derive reasonable abduction. 
In this sense, a natural rule of verbalizing the query would be adding a discourse connective ``after'' to convert the query to ``After PersonX gets tired of it, PersonX goes skydiving''. 
However, the verbalized query may still be ambiguous without additional context. 
To make the verbalized context more informative and human-understandable, we take advantage of Large Language Models (i.e., ChatGPT) to add additional context to compose the query to a narrative.

\paragraph{Relation Verbalization}

We use conversion rules and pre-defined templates to compose questions based on the relations in the queries.
Based on the definition of each commonsense relation \cite{DBLP:conf/aaai/HwangBBDSBC21}, we use the templates in \Cref{tab:prompt_relation} to verbalize each relation.
In terms of complex queries, we use the conversion rules in \Cref{tab:prompt_complex_relation} to convert the query to a question.

\paragraph{Person Names}

To make the context more natural, we replace PersonX, PersonY, PersonZ in the context to names randomly sampled from the 2021 public US social security application
name registry\footnote{{https://catalog.data.gov/dataset/baby-names-from-social-security-card-applications-national-data}}.

\begin{table}[t]
\small
\centering
\newcolumntype{C}[1]{>{\centering\arraybackslash}p{#1}}
\begin{tabular}{m{1.8em}|m{20em}}
\toprule
Query Type & Question Template \\
\midrule
   2i & What event or state is both Prompt(r1) [V1] and also prompt(r2) [V2]? \\
   3i & What event or state is both Prompt(r1) [V1], Prompt(r2) [V2], and also Prompt(r2) [V2]? \\
   2p & What event or state is Prompt(r1) \{Prompt(r2) [V1]\}? \\
   ip & What event or state is prompt(r3) \{both prompt(r1) [V1], and also prompt(r2) [V2] \}? \\
   pi & What event or state is both prompt(r1) \{prompt(r3) [V3]\}, and also prompt(r2) [V2]?\\
\bottomrule
\end{tabular}
\caption{Templates for verbalizing one-hop relations.}
\label{tab:prompt_relation}
\end{table}

\begin{table}[t]
\small
\centering
\newcolumntype{C}[1]{>{\centering\arraybackslash}p{#1}}
\begin{tabular}{m{5em}|m{16em}}
\toprule
Relation & Prompt Template \\
\midrule
   xIntent &  the intention of PersonX before\\
    xNeed &  what PersonX needed to do before\\
    xWant &  what PersonX wants to do after\\
    xEffect &  the effect on PersonX after\\
    xReact &  what PersonX feels after\\
    xAttr &  what PersonX is seen as given\\
    oEffect &  the effect on PersonY after\\
    oReact &  what PersonY feels after\\
    oWant &  what PersonY wants to do after\\
    HinderedBy &  what hindered\\
    isAfter &  what happens before\\
    isBefore &  what happens after\\
\bottomrule
\end{tabular}
\caption{Templates for verbalizing relations in complex queries. }
\label{tab:prompt_complex_relation}
\end{table}

\subsection{Human Annotation}
\label{sec:appx_human_annotation}

We introduce the details of the annotation process in this subsection.

\paragraph{Worker Selection} 
We have a qualification test to select eligible workers for the main task.
We prepare six pre-selected 2i queries of different types, including (negated) common effect, (negated) common cause, common attribute, and abduction.
Only Master annotators are eligible for participating the qualification.
We compare the pair-wise annotation accuracy between each annotator and the gold answer annotated by the authors of the paper, and select those who have at least 85\% agreement as qualified workers.
After selection, we pick 53 worker out of 120 participants in the qualification round.

\begin{figure*}[t]
    \centering
    \includegraphics[width=1.0\linewidth]{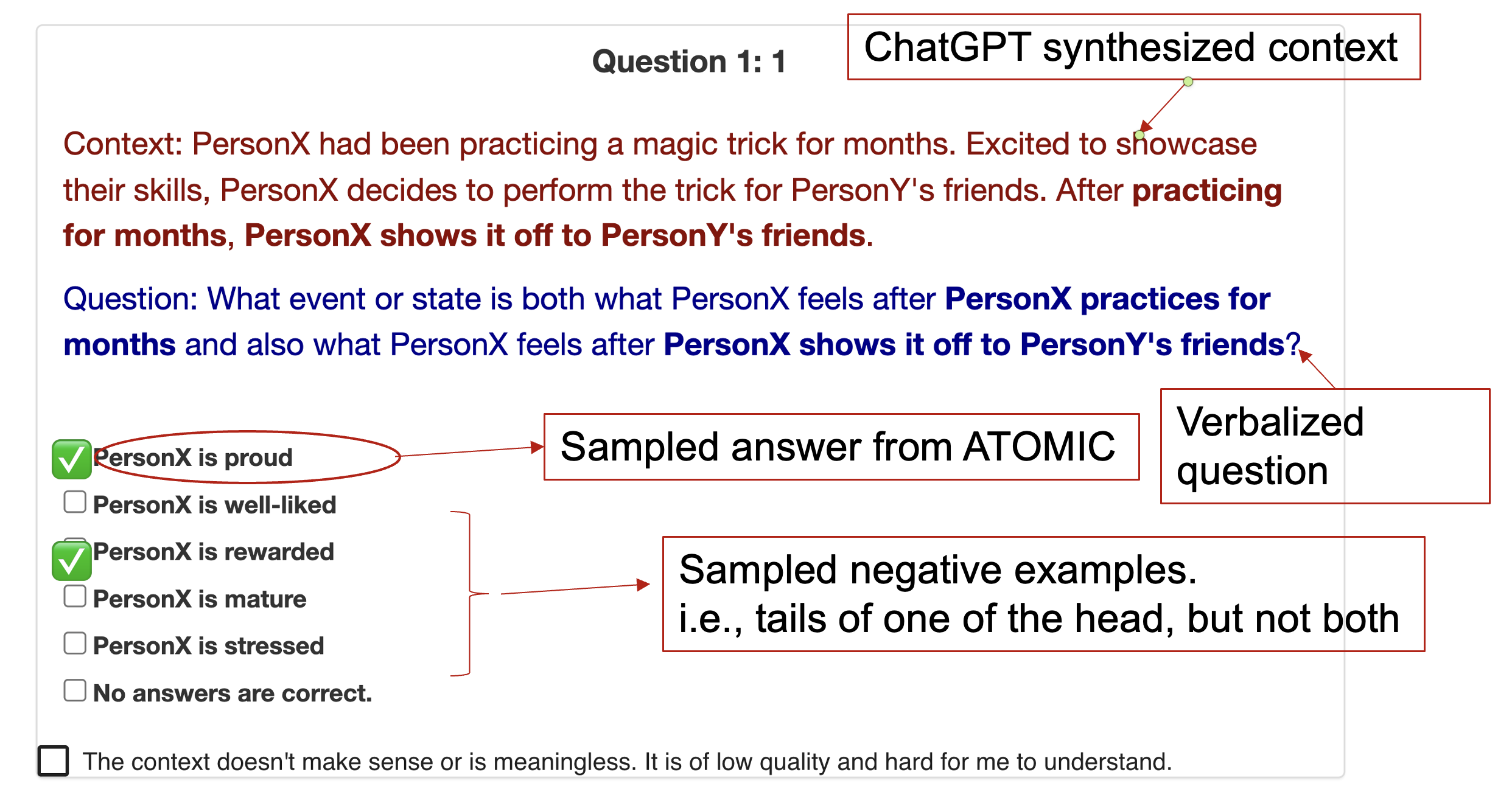}
    % \vspace{-0.12in}
    \caption{Annotation interface. }
    \label{fig:annotation_interface}
    % \vspace{-0.2in}
\end{figure*}

\paragraph{Annotation Interface}

A snapshot of the annotation interface is presented at \Cref{fig:annotation_interface}. 
In addition, we have provided comprehensive instructions along with detailed examples to guide the annotators throughout the annotation process. 
To ensure their understanding, we require annotators to confirm that they have thoroughly read the instructions by checking a checkbox before the annotation task.
We also manually checked the performance of the annotators along with the annotation process and gave feedbacks based on common errors.
For example, typical errors include mistakenly regard the one-hop answer as correct instead of fully considering the multi-hop context.

\paragraph{Post-processing}

To aggregate the annotation result, we randomly sample one option that is labeled as plausible by majority voting as the final positive answer, and sample three negative options and distractors.
If there are no options labeled as plausible, then the correct answer is ``None of the answers are correct''.
If there are less than three options labeled as negative, we manually add one or two negative examples to match the number.
To improve the quality, after crowdsourcing, the authors of this paper manually checked the QA pairs with an IAA lower than 0.6, and resolve the disagreements manually.

% \Cref{tab:stat} presents the statistics of the training and evaluation set.

\begin{table*}[t]
\small
\centering
\newcolumntype{C}[1]{>{\centering\arraybackslash}p{#1}}
\begin{tabular}{m{7em}|m{40em}}
\toprule
Model & Prompt \\
\midrule
\tabincell{m{7em}}{Llama2, Flan-T5\\ ChatGPT, GPT-4} & \tabincell{m{40em}}{Answer this commonsense reasoning question, where you are supposed to handle a multiple-chioce question answering task to select the correct answer. Select one correct answer from A to E.\slashn \\ \\ Context: [Context] Question: [Question] A: [Option A]. B: [Option B]. C: [Option C]. D: [Option D]. E: [Option E]. \slashn \\ \\ Answer:
}  \\
\midrule
 UnifiedQA & \tabincell{m{40em}}{[Question] \slashn  \\ (a): [Option A] (b) [Option B] (c) [Option C] (d) [Option D] (e) [Option E] \slashn \\  {[Context]} }\\ 
 \midrule
 Vera & [Context] [Question] [Option] \\
 \midrule
 HyKAS, CAR & [Context] [Question] [Option] \\
\bottomrule
\end{tabular}
\caption{Prompt templates for multiple-choice question answering.}
\label{tab:prompt_mcqa}
\end{table*}

\begin{table*}[t]
\small
\centering
\newcolumntype{C}[1]{>{\centering\arraybackslash}p{#1}}
\begin{tabular}{m{9em}|m{39em}}
\toprule
Model & Prompt \\
\midrule
Llama2 (zero-shot) & \tabincell{m{39em}}{[System\_Message] = As an expert in commonsense reasoning, your task is to provide a concise response to a question based on the given context. The question focuses on studying the causes, effects, or attributes of personas related to the given context. Answer shortly with no more than 5 words.\\ \\ <s>[INST] {<}{<}SYS{>}{>}\slashn {[System\_Message]} \slashn{<}{<}/SYS{>}{>}\slashn\slashn [Context] [Question] [/INST]}\\
\midrule
Llama2 (fine-tuned) & <|im\_start|>question\slashn [Context] [Question] <|im\_end|>\slashn <|im\_start|>answer\slashn [Answer] \\
\midrule
GPT-2 & \tabincell{m{39em}}{2i: [V1] [V2] [r1] [r2] [GEN] [Answer] \\ 3i: [V1] [V2] [V3] [r1] [r2] [r3] [GEN] [Answer]\\ 2p: [V1] [r1] [r2] [GEN] [Answer]}\\
\bottomrule
\end{tabular}
\caption{Prompts for fine-tuning generative commonsense inference models.}
\label{tab:prompt_comet}
\end{table*}

\section{Additional Details of Experiments}\label{sec:appx_exp}

% \section{Zero-shot CSQA}\label{sec:appx_zsqa}

% \subsection{Zero-shot CSQA}\label{sec:appx_zsqa_verbalization}

% \subsection{Negative Sampling}\label{sec:appx_zsqa_neg_sampling}

\subsection{Implementation Details of the Question Answering Models}

% \paragraph{Hyperparameters}

% \paragraph{Prompt Templates}

We follow the pipeline in HyKAS~\cite{DBLP:conf/aaai/MaIFBNO21} and CAR~\cite{wang-etal-2023-car}
% , and train our QA model by fine-tuning a pre-trained Masked Language Model (MLM) using the Marginal Ranking (MR) loss.
Let $C$ represent the original context, which is the head entity for 1p triple and the verbalized context for complex queries,
$Q$ represent the question verbalized from the anchor relations, and $(A_1, A_2,...)$ be the list of options.
We first concatenate $C$, $Q$, and an answer option $A_i$ together via natural language prompts following the order of ``$C$ $Q$ $A_i$'' to generate input sequences $(T_1,T_2,...)$.
% For example, the synthesized question with its correct answer in Figure~\ref{fig:CAR_overview} will be transformed as: ``PersonX arrives at the bar, as a result, PersonX want to, relax himself.''
We then repeatedly mask out one token at a time to calculate the masked language modeling loss. 
% The final MLM score for an input sequence $T \in \{T_1,T_2,...\}$ with $n$ tokens is:
\begin{equation}
\label{eq:MLM_score}
    \mathcal{S}(T)=-\frac{1}{n}\sum^n_{i=1}\log P(t_i | ...,t_{i-1},t_{i+1},...)
\end{equation}

% After calculating the scores $S_1, S_2,...$ for all answer candidates $A_1, A_2, ...$, 
We then compute the marginal ranking loss based on Equation~\ref{eq:margin_ranking_loss}, where $\eta$ represents the margin and $y$ is the index of the correct answer.
\begin{equation}
\label{eq:margin_ranking_loss}
\mathcal{L}=\frac{1}{m}\sum^m_{i=1, i\neq y}\max (0, \eta-S_y+S_i)
\end{equation}

% During evaluation, we use the same scoring procedure to assign a score to each option and select the one whose concatenated sentence achieves the lowest score as the model's prediction.
We train the DeBERTa QA model for 1 epoch with a learning rate of 5e-6 and a linear learning rate decay.
The checkpoint that yields the best performance on the synthetic validation set in CAR~\cite{wang-etal-2023-car} or HyKAS~\cite{DBLP:conf/aaai/MaIFBNO21} is selected as the final model.
During evaluating, we select the option that yields the lowest score as the final prediction.

We provide the prompt templates for each model in \Cref{tab:prompt_mcqa}.

\subsection{Implementation Details of Generative Commonsense Inference Models}\label{sec:appx_generative_experiments}

The training and evaluation of GPT2-based model is based on the paradigm defined in COMET~\cite{DBLP:conf/acl/BosselutRSMCC19}.
The input of one-hop ATOMIC triples is serialized to ``$h$ $r$'' and the expected output is $t$, where ($h$, $r$, $t$) forms a triple in the CSKG.
The input of 2p queries, ($h$, $r_1$, $V$) and ($V$, $r_2$, $V_?$), are serialized as ``$h$ $r_1$ $r_2$'' and the expected output is $V_?$.
The input of 2i queries, which includes ($h_1$, $r_1$, $V_?$) and ($h_2$, $r_2$, $V_?$), is serialized as ``$h_1$ $h_2$ $r_1$ $r_2$'' with the expected output as $V_?$.
All models are fine-tuned for 3 epochs with a batch size of 32, a learning rate of 1e-5, a linear learning rate decay.
The last checkpoint is taken as the final model.

For Llama2, we follow the standard instruction tuning procedure and use the pipeline provided by \citet{DBLP:journals/corr/abs-2311-16079}.
We train the model with a batch size of 32, learning rate of 1e-5, and linear learning rate decay.
We take the final checkpoint as our model to make prediction.

The whole list of prompt templates that we use is presented in \Cref{tab:prompt_comet}.

\section{Additional Analysis}\label{sec:appx_analysis}

\paragraph{Differences from ParaCOMET and COMET-M}
In ParaCOMET, the task involves providing a narrative as input, requiring the model to determine the commonsense causes or effects of a specific sentence within the context. 
To generate training data, a single-hop COMET model fine-tuned on ATOMIC is employed to create synthetic inferences. 
These inferences are generated solely based on the target sentence and the desired relation, without accessing the whole context. 
The resulting one-hop synthetic inferences are then utilized as distant supervision signals during the fine-tuning process for ParaCOMET.

COMET-M utilizes a context consisting of a sentence containing multiple events. 
Unlike from a sentence level, COMET-M focuses on generating commonsense inferences based on a specific event within the sentence. T
his fine-grained approach enables more precise and detailed commonsense reasoning.

In contrast, our complex commonsense reasoning benchmark introduces additional complexities compared to ParaCOMET and COMET-M. 
Besides the complex structures in the context that involves multiple events,
the desired relation or question involves multi-hop reasoning as well.  
For instance, rather than focusing on the cause of a single sentence or event, \dataname{} explores questions related to common causes, effects, attributions of multiple events, and two-hop inferences. 
This distinctive formulation sets our work apart and poses a greater challenge for LLMs to effectively reason and provide accurate responses.

\paragraph{Results of the Ablations}

We present the results of the ablation study in \Cref{tab:ablation}.

% \paragraph{The correctness of sampled queries}

\subsection{Difficulty of Different Query Types}

The results in \Cref{tab:mcqa_main_results} showed that performance varied depending on the evaluation query types.
Interestingly, pi queries exhibited a significantly higher success rate compared to other query types, particularly ip queries, considering both pi and ip involve a single free variable and both intersection and projection operations.
We present two perspectives to explain this phenomenon. 
First, the limited availability of sampled pi queries restricts the diversity of the data. 
Out of all the queries sampled from the development set of ATOMIC$_{20}^{20}$, only 4k are pi queries, while there are 12k ip queries and 598k 2i queries. 
This paucity of pi queries contributes to a lack of variety. 
Moreover, within these 4k pi queries, the number of unique answers is limited to 459, indicating a limited range of possible responses. 
As a result, models fine-tuned on ATOMIC can generate answers to pi queries more easily, given that most of them consist of nodes with high degrees.
Second, the chances of the sampled answer is actually the correct answer to pi queries (67.8\%) is significantly higher than other query types (e.g., 47.2\% for ip). 
This is also a result of the first reason, as the answers to the sampled queries are limited to nodes with high degrees, which are usually events with a broad meaning such as ``PersonX gets better''.

% In all, despite that the query structure itself is more complicated, the reasoning difficulty is not that hard compared to other query types due to the above two reasons.

\paragraph{Discussions on Further Applications of Complex Queries}

Intuitively, 2i queries can represent various scenarios such as common attribution, common effect, common cause, and abduction (when one relation pertains to effects and the other relates to cause), depending on the types of relations involved in the query.
Besides, complex logical queries, particularly those involving intersection operations, are relevant to defeasible reasoning~\cite{DBLP:conf/emnlp/RudingerSHBFBSC20}, where inferences can be weakened given new evidence. 
In the one-hop setting, tails are annotated in a context-free manner, considering only the most general cases. 
However, in intersection-based queries like 2i and 3i, additional anchor entities and relations act as specific constraints, narrowing down the inferences to a particular scope while disregarding other commonsense inferences in the context-free scenario. 
For instance, in the example from \Cref{fig:Introduction_demo}, other potential tails for (PersonX goes skydiving, xIntent) could include overcoming fear, seeking enjoyment, or achieving a personal milestone. Nevertheless, when constrained by another query (PersonX gets tired of it, xWant), the intentions related to fear, enjoyment, and fulfillment are weakened, and only the correct inference of ``finding new things to do'' remains.

\begin{table*}[t]
\scriptsize
\centering
\newcolumntype{C}[1]{>{\centering\arraybackslash}p{#1}}
\begin{tabular}{@{}l@{}|m{18em}|m{18em}|m{10em}|m{10em}}
\toprule
Type & Context & Question & COMET & \dataname{}-COMET \\
\midrule
2p & Ezra updates Ezra's resume (\textcolor{entitycolor}{V1}) & What event or state is \textcolor{relation1color}{the intention of Ezra before} \textcolor{relation2color}{the intention of Ezra before} \textcolor{entitycolor}{V1}? & \tabincell{l}{get a new job \xmark \\(one-hop correct)} & be financially independent \cmark \\
\midrule
\tabincell{m{2em}}{2i-\\neg} & \textcolor{entitycolor}{Every day, Benjamin goes to work diligently (V1)}, never missing a day. They are dedicated and committed to their job. In particular, \textcolor{entitycolor}{Benjamin doesn't work hard on it (V2)} and instead takes a more relaxed approach, focusing on maintaining a healthy work-life balance. & What event or state is both \textcolor{relation1color}{the effect on Benjamin after} \textcolor{entitycolor}{Benjamin go to work every day (V1)} and also \textcolor{relation2color}{what hindered} \textcolor{entitycolor}{Benjamin work hard on it (V2)}? & {Benjamin is sick {\fontfamily{cyklop}\selectfont \textit{?}}\qquad  (Not perfect as Benjamin is trying to keep a work-life balance instead of having a sick leave)}  & Benjamin gets tired from working hard \cmark \\
\midrule
2i & \textcolor{entitycolor}{Chloe is known for being hardworking (V1)} and dedicated. As a result, \textcolor{entitycolor}{Chloe leads a good life (V2)}. & What event or state is both \textcolor{relation1color}{the effect on Chloe after} \textcolor{entitycolor}{Chloe is hardworking (V1)} and also \textcolor{relation2color}{what Chloe wants to do after} \textcolor{entitycolor}{Chloe leads a good life (V2)}? & to have a good life {\fontfamily{cyklop}\selectfont \textit{?}}   (No inferential gap) & to have success in life {\fontfamily{cyklop}\selectfont \textit{?}} (No inferential gap) \\
\midrule
ip & After \textcolor{entitycolor}{looking for a new car (V1)}, \textcolor{entitycolor}{Lydia is driving to school (V2)}. & What event or state is \textcolor{relation2color}{what Lydia needed to do before} the event that is both \textcolor{relation1color}{what Lydia wants to do after} \textcolor{entitycolor}{Lydia is looking for a new car (V1)}, and also \textcolor{relation1color}{what Lydia needed to do before} \textcolor{entitycolor}{Lydia is driving to school (V2)}? & None \xmark & take a car for test drive \cmark \\
\bottomrule
\end{tabular}
\caption{Error analysis of generated inferences on the evaluation set of \dataname{}. We present the generations of COMET-Llama-7b and \dataname{}-Llama-7b fine-tuned on all queries.}
\label{tab:cases}
\end{table*}

\section{Error Analysis}\label{sec:appx_error_analysis}

We present some error cases in \Cref{tab:cases}. 
In general, a common error in both projection and intersection queries is that the generated answer can be only the one-hop answer instead of the correct answer that is multi-hop.
For example, in the 2p case, ``get a new job'' is a direct intention of someone who updates his or her resume. However, the 2p query asks about the intention of the intention, which requires inducing the intention behind ``get a new job''. In this sense, ``to be financially independent'' is more plausible inference.
In the case of 2i queries, the error lies in the absence of inferential gaps between the context, where the generated answers become paraphrases of the events rather than being the result by any anchor entity. In the case of ip, a common error for one-hop COMET is the generation of ``None'' for complex cases, indicating a deficiency in multi-hop reasoning capabilities.

\end{document}